\documentclass[10pt,twocolumn,letterpaper]{article}

\usepackage[pagenumbers]{cvpr} 

\definecolor{cvprblue}{rgb}{0.21,0.49,0.74}
\usepackage[pagebackref,breaklinks,colorlinks,allcolors=cvprblue]{hyperref}

\def\paperID{594} 
\def\confName{CVPR}
\def\confYear{2026}

\title{WildCap: Facial Albedo Capture in the Wild via Hybrid Inverse Rendering}

\author{
Yuxuan Han\textsuperscript{1}\quad
Xin Ming\textsuperscript{1}\quad
Tianxiao Li\textsuperscript{1}\quad
Zhuofan Shen\textsuperscript{1}\quad
Qixuan Zhang\textsuperscript{2,3}\quad
Lan Xu\textsuperscript{2}\quad
Feng Xu\textsuperscript{1}
\and
\textsuperscript{1}School of Software and BNRist, Tsinghua University 
\textsuperscript{2}ShanghaiTech University
\textsuperscript{3}Deemos Technology
}

\begin{document}

\twocolumn[{%
\renewcommand\twocolumn[1][]{#1}%
\maketitle

\begin{center}
    \centering
    \captionsetup{type=figure}
    \includegraphics[width=\textwidth]{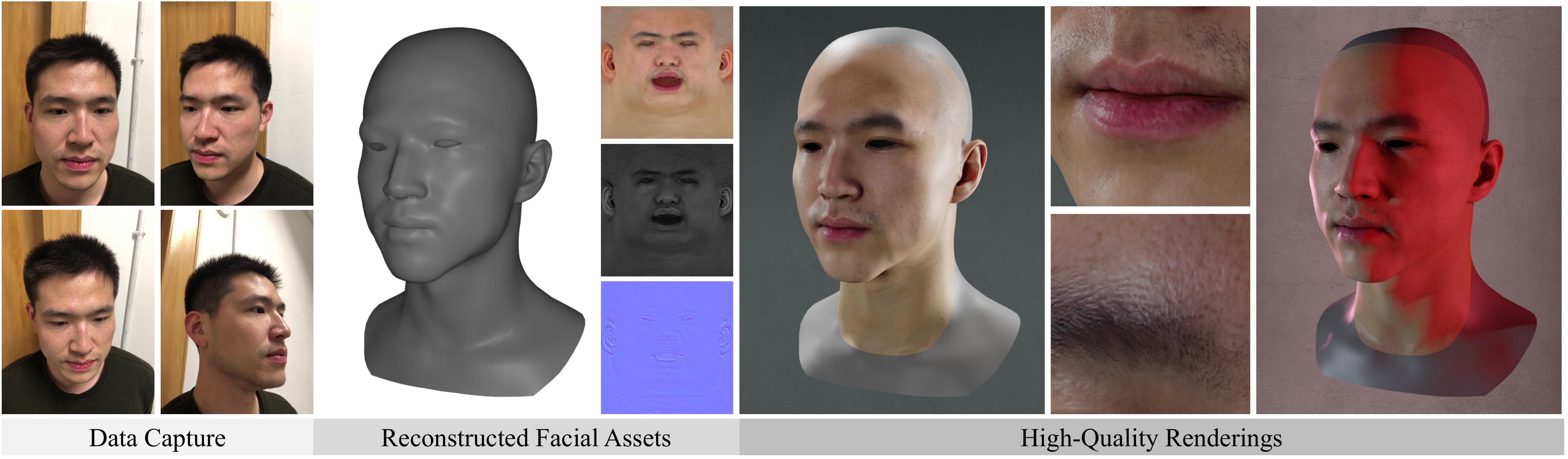}
    \captionof{figure}{
    \textbf{WildCap:} Given a smartphone video captured in the wild (4 frames shown above), our method reconstructs a high-quality diffuse albedo map and predicts other reflectance maps from it, which can be exported to graphics engines for photo-realistic rendering. }
    \label{Fig:teaser}
\end{center}%
}]

\begin{abstract}
    Existing methods achieve high-quality facial albedo capture under controllable lighting, which increases capture cost and limits usability. We propose WildCap, a novel method for high-quality facial albedo capture from a smartphone video recorded in the wild. To disentangle high-quality albedo from complex lighting effects in in-the-wild captures, we propose a novel hybrid inverse rendering framework. We first apply a data-driven method, i.e., SwitchLight, to convert the captured images into more constrained conditions and then adopt model-based inverse rendering. However, unavoidable local artifacts in network predictions, such as shadow-baking, are non-physical and thus hinder accurate inverse rendering of lighting and material. To address this, we propose a novel texel grid lighting model to explain non-physical effects as clean albedo illuminated by local physical lighting. During optimization, we jointly sample a diffusion prior for the albedo map and optimize the lighting, effectively resolving scale ambiguity between local lights and albedo. Other reflectance maps are then predicted from the albedo. Our method achieves significantly better results than prior arts in the same capture setup, closing the quality gap between in-the-wild and controllable recordings by a large margin. Our code is released \href{https://yxuhan.github.io/WildCap/index.html}{\textcolor{magenta}{here}}.
\end{abstract}

\section{Introduction}
This paper focuses on facial albedo capture, where the goal is to reconstruct facial diffuse albedo maps from images.
As the core step of cloning human beings into the digital world, this problem has been extensively studied in the past two decades~\cite{klehm2015recent}.
Although production-level results are demonstrated~\cite{alexander2009digital,alexander2013digital}, previous success relies on the assumption that the scene lighting is controllable, ranging from using the Light Stage in a high-end setup~\cite{ma2007rapid,ghosh2011multiview,debevec2000acquiring} to the smartphone flashlight in a low-cost setup~\cite{han2024cora,han2025dora,azinovic2023high}.
However, this reliance on the scene lighting inevitably increases capture cost and limits usability.
Thus, we ask if high-quality facial albedo capture can be achieved from images captured in the wild, without ANY assumptions about the scene lighting.

In the literature, model-based methods optimize lighting and facial reflectance maps to match the captured images via differentiable rendering~\cite{dib2021practical,bharadwaj2023flare,xu2024monocular}.
Although these methods work well in simple cases with low-frequency lighting, they struggle in most in-the-wild cases with complex light transport, as the optimization is unstable and inherently ill-posed.
On the other hand, data-driven methods learn a neural network to predict facial reflectance components directly from images~\cite{wang2020single,pandey2021total,kim2024switchlight,yeh2022learning,chaturvedi2025synthlight}.
Through large-scale training, these methods are robust to in-the-wild faces.
However, it is challenging for a neural network to fully understand the physical law of light transport, thus inevitably leaving artifacts like shadow-baking in their predictions as shown in Figure~\ref{Fig:cmp_ablat} (b).
Therefore, it still remains a challenge to reconstruct a high-quality facial diffuse albedo map from images captured in the wild.

In this paper, we aim to fill the quality gap between in-the-wild methods and the methods under controllable lighting.
We propose WildCap, a novel approach that reconstructs a high-quality facial diffuse albedo map from multi-view images captured in the wild by a smartphone.
To this end, a hybrid inverse rendering approach is proposed to combine the model-based and data-driven methods.
Specifically, we first apply a data-driven method, \emph{i.e.}, SwitchLight~\cite{kim2024switchlight}, to predict the diffuse albedo image for each input view.
As data-driven predictions are \emph{not perfect} (\emph{e.g.}, they might bake some lighting effects such as shadow), we then treat these predicted diffuse albedo images as real images (but captured under less challenging lighting), and factor them into the lighting and \emph{clean} diffuse albedo map via differentiable rendering.
This way, the model-based optimization becomes more stable, as the data-driven method has already converted potentially complex in-the-wild lighting conditions into simpler and more constrained ones. 

Although conceptually simple, achieving this is not easy.
Unlike the original captured images, the predicted diffuse albedo images are generated by a neural network, \emph{i.e.}, SwitchLight~\cite{kim2024switchlight}, rather than physical light sources in the real world.
Thus, conventional physics-based lighting models cannot explain the non-physical baking artifacts on these predicted diffuse albedo images as lighting effects.
To address this, we propose the texel grid lighting model, a novel non-physical but more expressive lighting representation.
Specifically, we model lighting as a 2D grid with Spherical Harmonics (SH)~\cite{ramamoorthi2001efficient} in the UV space, which compactly represent global variant and local smooth lighting.
For each UV texel, we query its SH parameters from this grid via bilinear interpolation.
This way, different facial regions are modeled with different SH lighting. 
In turn, we gain sufficient expressive capacity to model these non-physical effects and further clean up baking artifacts in the predicted diffuse albedo images.

However, as our texel grid lighting model has increased expressive power, the optimization becomes more ill-posed.
Without regularization, we cannot ensure the baking artifacts are decomposed into a valid diffuse albedo map illuminated by a local light as we expect.
To this end, we propose to estimate the albedo within the prior distribution of high-quality and valid ones.
Inspired by DoRA~\cite{han2025dora}, we learn a patch-level diffusion prior for facial diffuse albedo maps over Light Stage scans.
We then apply the diffusion posterior sampling technique~\cite{chung2022diffusion} to steer this patch-level diffusion model to generate a full-resolution diffuse albedo map that best matches the observations, \emph{i.e.}, the predicted diffuse albedo images.
By jointly optimizing the texel grid lighting model and sampling the diffusion prior, we effectively reconstruct a high-quality diffuse albedo map from in-the-wild captures.
Other reflectance maps, \emph{i.e.}, specular and normal maps, are then predicted from the albedo.
In conclusion, our main contributions include:
\begin{itemize}
    \item A novel hybrid inverse rendering method for facial albedo capture that closes the quality gap between in-the-wild and controllable recordings by a large margin.
    \item A texel grid lighting model to represent non-physical lighting effects in network-predicted images.
    \item A scheme for jointly optimizing the texel grid lighting model and sampling the diffusion prior for high-quality facial diffuse albedo map reconstruction.
\end{itemize}

Our code is released to foster future research.
We hope our method can serve as a handheld Light Stage to scan everyday users to enter the digital world.

\section{Related Works}
\subsection{Inverse Rendering}
The goal of inverse rendering is to reconstruct the geometry, reflectance, and scene lighting from images.
This problem has been extensively studied in recent years.
Previous model-based methods typically use neural fields~\cite{mildenhall2020nerf,xie2022neural,kerbl3Dgaussians} to represent geometry and reflectance, and integrate the physics-based lighting model into the optimization~\cite{physg2021,zhang2021nerfactor,Munkberg_2022_CVPR,hasselgren2022nvdiffrecmc,Li:2024:TensoSDF,liu2023nero,boss2021nerd,boss2021neuralpil,sun2023neural,zhang2022invrender,yao2022neilf}. 
Many of them adopt an environment map to represent scene lighting.
They further apply the Spherical Gaussian~\cite{zhang2022invrender,physg2021} or pre-integrated lighting~\cite{Munkberg_2022_CVPR,boss2021neuralpil} for efficient inverse rendering.
To reconstruct clean reflectance maps without baking artifacts, some works introduce differentiable ray tracing~\cite{Mitsuba3} to optimization~\cite{sun2023neural,hasselgren2022nvdiffrecmc}.
Other works use neural fields as a cache to simulate global light transport effects~\cite{zhang2021nerfactor,zhang2022invrender,yao2022neilf}.
More recently, data-driven methods propose to train a neural network to predict reflectance components directly from images~\cite{chen2025intrinsicanything,hong2024supermatphysicallyconsistentpbr,litman2025materialfusion}.
These neural networks can serve as a strong prior to regularize the model-based method~\cite{litman2025materialfusion,chen2025intrinsicanything}.
Nonetheless, one of the core challenges in this problem today is reconstructing clean reflectance maps without baking lighting effects such as shadows.

\subsection{Facial Appearance Capture}
Facial appearance capture is an application of the inverse rendering problem discussed before.
Previous works achieve high-quality results on images captured under controllable lighting. 
High-end methods build professional apparatus~\cite {debevec2012light,riviere2020single,lattas2022practical,zhang2022videodriven} in studios for data capture.
Some works reconstruct facial reflectance maps from dense One-Light-At-a-Time (OLAT) images~\cite{debevec2000acquiring,weyrich2006analysis} or special lighting patterns~\cite{ma2007rapid,ghosh2011multiview}.
Other works propose to capture facial appearance in a single-shot setup~\cite{riviere2020single,xu2022improved,gotardo2018practical}.
On the other hand, low-cost methods exploit smartphone flashlights~\cite{azinovic2023high,han2024cora,han2025dora} or sunlight~\cite{wang2023sunstage} for controllable data capture.
However, their reliance on scene lighting increases capture cost and limits usability.

To address this, recent works propose capturing facial appearance from in-the-wild images.
A group of works reconstruct a relightable scan from a single face image by training on the Light Stage dataset~\cite{lattas2020avatarme,lattas2021avatarme++,lattas2023fitme,Paraperas_2023_ICCV,galanakis2025fitdiff,dib2021practical,Dib_2024_CVPR,han2023learning,smith2020morphable,huanglearning,rao20253dpr,prao2024lite2relight}.
To model lighting effects that are challenging to simulate with physics-based light transport, \emph{e.g.} external occlusions, DeFace~\cite{huanglearning} trains a network to segment the face into regions and apply different SH lightings to model each facial region.
Our method shares the same high-level spirit as DeFace; we apply non-physical texel grid lighting to represent what is impossible to model with physics-based lighting, \emph{i.e.}, the baking artifacts in network-predicted images.
However, our method is designed for high-quality diffuse albedo map reconstruction, while DeFace is limited to the statistical prior.

Other works consider a multi-view setup~\cite{xu2024monocular,FLAME:SiggraphAsia2017,rainer2023neural,zheng2023neuface,li2024uravatar}.
Compared to single-view methods, they obtain better results.
However, they still struggle in in-the-wild scenarios with complex lighting effects such as shadows.
Among them, \citet{rainer2023neural} applies a small MLP to directly model diffuse and specular shading, which has the potential to represent non-physical lighting effects in our scenario.
Instead, we apply the texel grid to model lighting.
Compared to the MLP, our grid representation is simpler to optimize within the diffusion posterior sampling framework.
In addition, to improve the results, we propose a novel hybrid method that combines data-driven delighting with model-based optimization.
On the one hand, our method inherits the robustness of data-driven methods.
On the other hand, model-based optimization can effectively remove baking artifacts from network predictions, resulting in high-quality and clean textures at 4K resolution.

\begin{figure*}[t]
    \centering
    \includegraphics[width=1\textwidth]{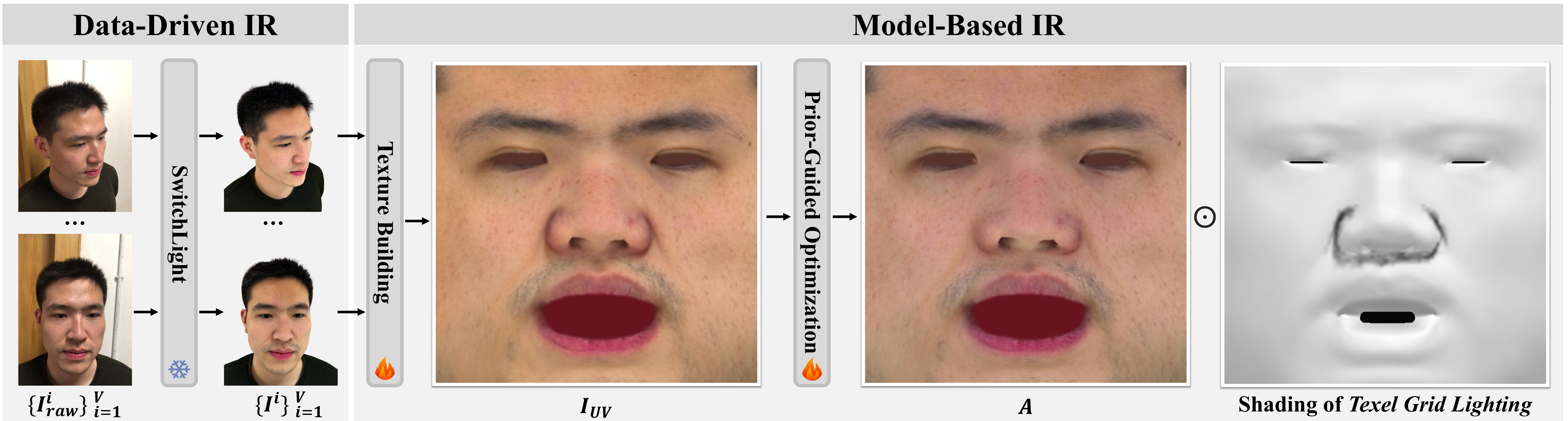}
    \caption{\textbf{Pipeline of our method.} We propose a novel hybrid inverse rendering (IR) framework for high-quality facial albedo capture in the wild.
    Given multi-view face images $\{I_{raw}^i\}_{i=1}^{V}$ captured by a smartphone, we apply a data-driven IR method, SwitchLight~\cite{kim2024switchlight}, to predict the diffuse albedo images $\{I^i\}_{i=1}^{V}$ for each view.
    Since SwitchLight is not perfect, we apply a model-based optimization in the UV space to explain its baking artifacts as lighting effects.
    This effectively removes the artifacts and produces a clean diffuse albedo $A$.
    }
    \label{Fig:ppl}
\end{figure*}

\begin{figure}[t]
    \centering
    \includegraphics[width=0.475\textwidth]{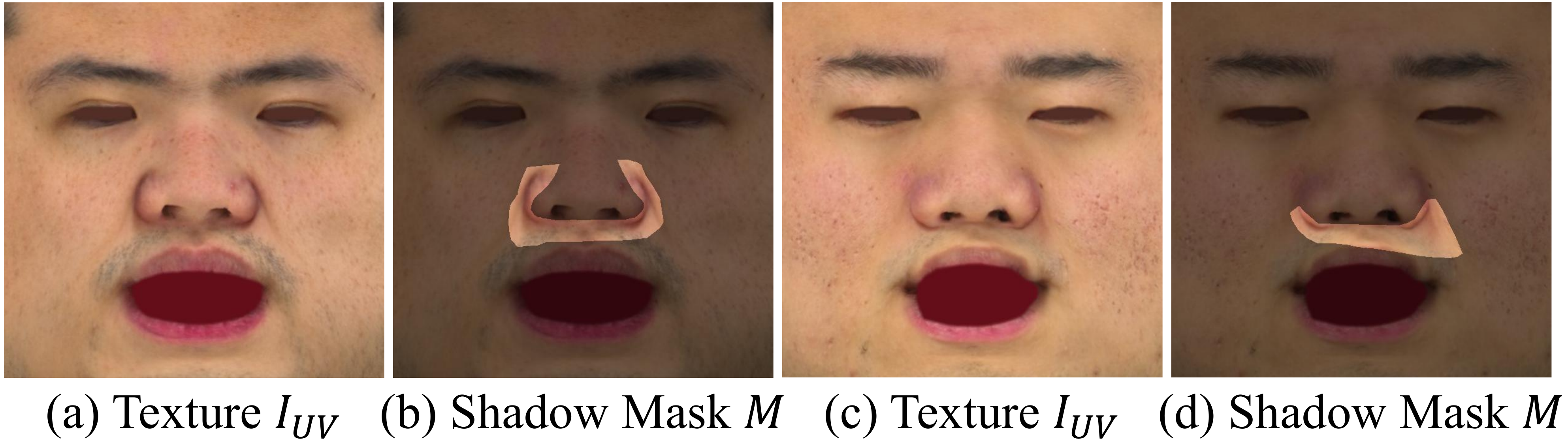}
    \caption{\textbf{Illustration of the shadow mask $M$.} We show the texture map in (a) and (c) and highlight the regions with shadow-baking artifacts in (b) and (d). In our texel grid lighting model, we apply a SH lighting grid to model the residual lighting inside $M$.}
    \label{Fig:shadow_mask}
\end{figure}

\section{Method}
In this Section, we first introduce our data capture and processing steps in Section~\ref{sec:method:data}.
To solve high-quality facial albedo maps from the captured data, we propose a hybrid inverse rendering framework in Section~\ref{sec:method:hybrid}.
Next, we detail the proposed texel grid lighting model (Section~\ref{sec:method:texel-grid}) and the optimization method (Section~\ref{sec:method:optimize}).

\subsection{Data Capture and Processing}\label{sec:method:data}
As shown in Figure~\ref{Fig:teaser}, we capture a smartphone video around the subject as input.
Unlike previous works~\cite{han2024cora,han2025dora,wang2023sunstage}, we have no assumptions about the scene lighting.
The capture takes about 30 seconds.
We empirically find that a non-professional user can keep still during the capture.
We uniformly sample 300 frames from the video and resize them to 960$\times$720 resolution.
We calibrate the camera parameters for each frame using COLMAP~\cite{schoenberger2016mvs,schoenberger2016sfm} and reconstruct a detailed mesh using 2DGS~\cite{Huang2DGS2024}.
We use Wrap3D~\cite{Faceform_Wrap2025} to register the ICT template~\cite{li2020learning,zhang2023hack} to the detailed mesh as our geometry $\mathcal{G}$.
We further sample $V=16$ frames $\{I_{raw}^i\}_{i=1}^{V}$ from all 300 frames according to sharpness for reflectance estimation.
To resolve the color ambiguity between albedo and lighting, we assume the skin tone of the captured subject is provided; this can be done either manually or automatically~\cite{feng2022towards,ren2023improving}.

\subsection{Hybrid Inverse Rendering}\label{sec:method:hybrid}
Given the captured images $\{I_{raw}^i\}_{i=1}^{V}$, geometry $\mathcal{G}$, and camera parameters, conventional model-based inverse rendering methods optimize the lighting and a set of reflectance maps so that the re-rendered images can match the captured ones~\cite{xu2024monocular,dib2021practical,bharadwaj2023flare}.
Despite working well in simple cases with low-frequency lighting, they struggle in most in-the-wild cases with complex light transport effects such as shadows.
The reason is that simulating this complex light transport makes the optimization process unstable and ill-posed.

On the other hand, data-driven inverse rendering methods adopt a neural network to directly predict the reflectance components from images~\cite{wang2020single,pandey2021total,kim2024switchlight,yeh2022learning,chaturvedi2025synthlight}.
These methods are robust to in-the-wild images, as they bypass explicit modeling of light transport.
However, as neural networks are not perfectly accurate, there are inevitably baking artifacts in their predictions as shown in Figure~\ref{Fig:cmp_ablat} (b).
Unfortunately, because the neural network is unexplainable, we can only accept these artifacts as it is.

In this paper, we propose a novel hybrid inverse rendering method to combine the best of two worlds.
As shown in Figure~\ref{Fig:ppl}, we first apply a data-driven method, SwitchLight~\cite{kim2024switchlight}, to predict the diffuse albedo of the captured images $\{I_{raw}^i\}_{i=1}^{V}$.
We select SwitchLight because it is the best publicly available method.
We denote these predicted diffuse albedo images as $\{I^i\}_{i=1}^{V}$.
Then, we apply a model-based inverse rendering method to $\{I^i\}_{i=1}^{V}$.
The rationale is to explain the baking artifacts in network predictions as lighting effects.
This way, we can obtain clean reflectance maps.
In practice, we focus on removing shadow-baking artifacts, as we find SwitchLight works well on facial specularity and other smooth lighting effects.

Specifically, we build a texture map $I_{UV}\in\mathbb{R}^{H\times W\times 3}$ from $\{I^i\}_{i=1}^{V}$ using the geometry $\mathcal{G}$ and camera parameters.
We then optimize the diffuse albedo map $A\in\mathbb{R}^{H\times W\times 3}$ and the lighting model $\Gamma_{\theta}$ to minimize the photometric loss in the UV space:
\begin{equation}
\label{eq:inv}
    \mathcal{L}_{pho}(A,\theta) = ||I_{UV} - \Gamma_{\theta}(A, N_c)||_2^2
\end{equation}
Here, $N_c\in\mathbb{R}^{H\times W\times 3}$ is the coarse normal map computed from $\mathcal{G}$.
Note that we assume faces are Lambertian surfaces, as we empirically find specularity in the captured images $\{I_{raw}^i\}_{i=1}^V$ is well removed by SwitchLight.
As demonstrated by previous works~\cite{li2021tofu,zhang2023dreamface,Li2020DynamicFA}, other reflectance maps, such as specular albedo $S\in\mathbb{R}^{H\times W}$ and detailed normal maps $N_{d}\in\mathbb{R}^{H\times W\times 3}$, can be inferred from $A$ with high quality.
Next, we introduce the lighting model $\Gamma_{\theta}$ and how we minimize Eq.~\eqref{eq:inv}.

\subsection{Texel Grid Lighting Model}\label{sec:method:texel-grid}
\begin{figure*}[t]
    \centering
    \includegraphics[width=1\textwidth]{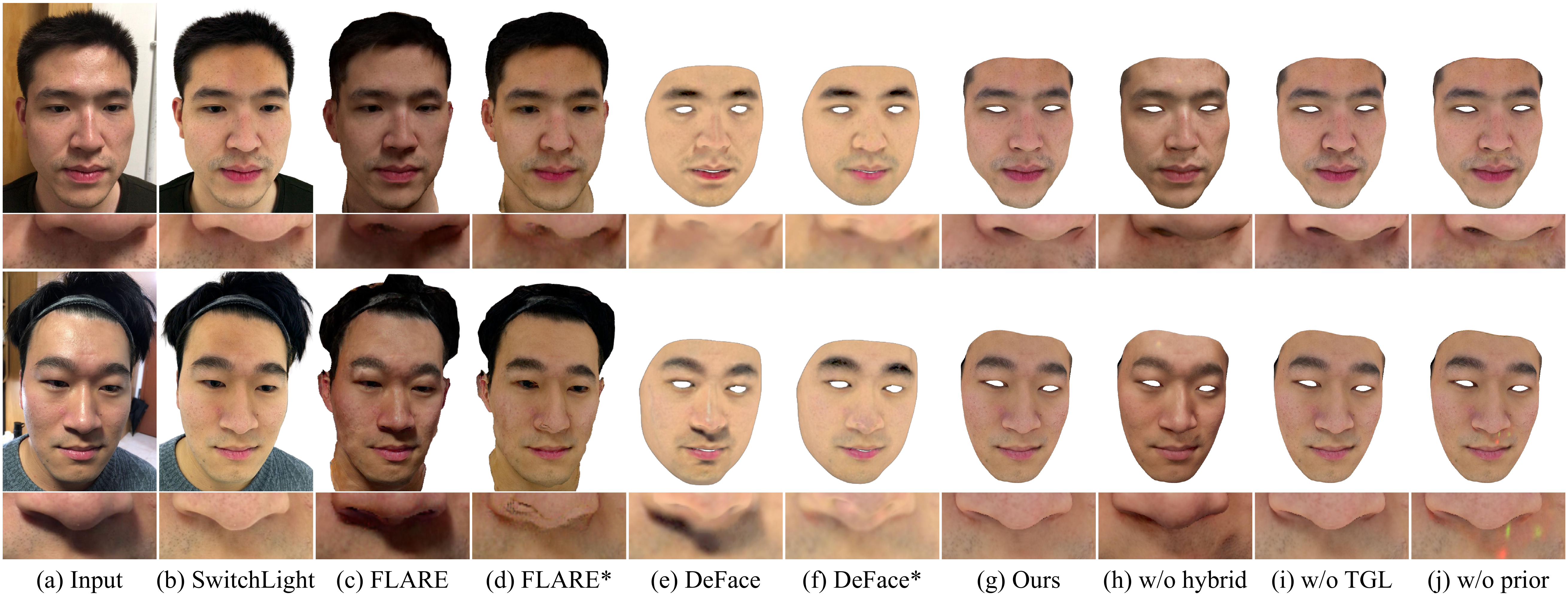}
    \caption{\textbf{Qualitative comparison and ablation study on diffuse albedo reconstruction.} 
    }
    \label{Fig:cmp_ablat}
\end{figure*}

To model lighting, previous works typically apply an environment map.
They further use the SH approximation~\cite{ramamoorthi2001efficient} to improve rendering efficiency.
However, this physics-based representation does not work well in our case, as the texture map $I_{UV}$ is produced by a neural network rather than physical light sources in the real world.
As shown in Figure~\ref{Fig:cmp_ablat} (i), implementing $\Gamma_\theta$ as an SH lighting model works well in most facial regions, but cannot explain the non-physical shadow-baking artifacts as lighting effects.

Based on this observation, we propose the texel grid lighting model, a novel non-physical but more expressive lighting representation tailored to our hybrid inverse rendering framework.
Our core idea is to assign more lighting models to facial regions with shadow-baking artifacts.
In this way, the shadow-baking artifacts can be explained as a clean diffuse albedo illuminated by local dark lights.

To this end, we first compute a binary mask $M\in\mathbb{R}^{H\times W}$ indicating shadow-baking artifacts in the UV space.
As shown in Figure~\ref{Fig:shadow_mask}, we only expect $M$ to roughly segment out the shadow-baking artifacts, thus imposing limited efforts to obtain it.
By default, we manually create $M$ to ensure the best quality.
We also propose an automatic method for computing $M$ using existing open-sourced shadow-soften  methods~\cite{ponglertnapakorn2023difareli,ponglertnapakorn2025difarelidiffusionfacerelighting}, which yields nearly comparable results to the manually created one; see our \emph{supplementary material} for more details.

Our lighting model includes two parts modulated by $M$:
\emph{i)} a 2D grid $V\in\mathbb{R}^{\frac{H}{g}\times \frac{W}{g}\times N_c}$ with SH parameters to model lighting in facial regions with baking artifacts, 
and \emph{ii)} a global SH lighting $\gamma^g\in\mathbb{R}^{N_c}$ to model the whole face.
Here, we adopt 2-order SH and $N_c=27$; $g$ is the grid size.
Given a texel with UV coordinate $(u,v)$, we first query $V$ via bilinear interpolation:
\begin{equation}
    \gamma^V = {\rm interp}(u, v; V)
\end{equation}
Then, we compute the SH parameters $\gamma\in\mathbb{R}^{27}$ for shading as the combinatation of $\gamma^g$ and $\gamma^V$ modulated by $M$:
\begin{equation}
    \gamma = \gamma^g + \gamma^V \cdot M[u][v]
\end{equation}
Next, we compute the texel color $c\in\mathbb{R}^3$ via standard SH shading~\cite{ramamoorthi2001efficient}:
\begin{equation}
    c = \frac{a}{\pi}\cdot \sum_{l=0}^2\sum_{m=-l}^{l} B_{l}\cdot\gamma_{lm}\cdot Y_{lm}(\textbf{n})
\end{equation}
Here, $a=A[u][v]\in\mathbb{R}^3$ and $\textbf{n}=N_c[u][v]\in\mathbb{R}^3$ are the diffuse albedo and normal value of the texel located at $(u,v)$, $B_{l}\in\mathbb{R}$ are the SH coefficients of the Lambertian BRDF, and $Y_{lm}$ are the SH basis functions.

However, as the expressive power of lighting increases, the optimization becomes more under-constrained.
If optimized freely, due to the inherent ambiguity between the scale of lighting and albedo~\cite{ramamoorthi2001signal}, we cannot ensure converging to a valid reflectance map illuminated by a dark light.
Next, we detail how we optimize our texel grid lighting model to produce a high-quality diffuse albedo map.

\subsection{Optimization}\label{sec:method:optimize}
To address the optimization dilemma, we propose solving the diffuse albedo map $A$ within the distribution of high-quality and valid ones.
Inspired by DoRA~\cite{han2025dora}, we train a patch-level diffusion model over 48 Light Stage scans from an online store.
Then, we jointly sample $A$ from the diffusion model and optimize the texel grid lighting model $\Gamma_\theta$ to minimize Eq.~\eqref{eq:inv}.

\paragraph{Diffusion Prior Training}
We follow DoRA's method to train the diffusion prior $\epsilon$ at $64\times64$ resolution to model the distribution of reflectance patches cropped from 1K-resolution reflectance maps.
Once trained, given a Gaussian noise $x_T$, $\epsilon$ can gradually denoise it into a clean sample $x_0$\footnote{Similar to DoRA, our diffusion prior $\epsilon$ also takes UV coordinate map as a condition. We omit it here for clarity.}; in our case, $x_0$ is the concatenation of the 3-channel diffuse albedo patch, 3-channel detailed normal patch, and 1-channel specular albedo patch along the channel axis:
\begin{equation}
\label{eq:diffusion_reverse}
    x_{t-1}=\frac{1}{\sqrt{\alpha_t}}\cdot\left(x_t-\frac{1-\alpha_t}{\sqrt{1-\bar{\alpha}_t}} \cdot\epsilon\left(x_t, t\right)\right)+\sigma_t\cdot z
\end{equation}
Here, $z$ is a standard Gaussian noise, $\alpha_t$, $\bar{\alpha}_t$, and $\sigma_t$ are predefined constants~\cite{ho2020denoising}.
In addition, at time step $t$, we can obtain the estimation of the clean data point $\hat{x}_t$ via:
\begin{equation}
\label{eq:diff_extimate_x0}
\hat{x}_t=\dfrac{1}{\sqrt{\bar{\alpha}_t}}\cdot(x_t-\sqrt{1-\bar{\alpha}_t}\cdot\epsilon(x_t,t))
\end{equation}

Although Eq.~\eqref{eq:inv} enforces no explicit constraints over the specular albedo and the detailed normal, we still model them in the diffusion prior $\epsilon$ as we find that sampling them together with $A$ can also produce a high-quality specular albedo map $S$ and detailed normal map $N_d$, which avoid extra efforts to train a network to predict them from $A$.

\paragraph{Initialization}
Before optimization, we require initializing the lighting model $\Gamma_\theta$.
To this end, we first select a scan $x_0^{ref}$ from our Light Stage dataset that has the most similar skin tone to the captured subject.
We further adjust the diffuse albedo components in $x_0^{ref}$ to the target skin tone via color matching.
We optimize $\gamma^g$ to minimize the photometric loss between $I_{UV}$ and the SH-shaded $x_0^{ref}$.
We initialize all SH parameters in $V$ as $0$.

Instead of sampling the diffusion model from the Gaussian noise $x_T$ as DoRA, we start with a cleaner data sample $x_{T_{init}}$.
Specifically, we add $T_{init}=0.6\cdot T$ steps of noise to $x_0^{ref}$ to obtain $x_{T_{init}}$.
Compared to DoRA, our method requires fewer sampling steps without sacrificing quality.

\paragraph{Jointly Sample $A$ and Update $\Gamma_\theta$}
Although the diffusion prior $\epsilon$ is trained at the patch level, as demonstrated by DoRA, we can directly sample it at higher resolution.
Specifically, we apply diffusion posterior sampling~\cite{chung2022diffusion} to sample an $x_0$ at 1K resolution that can minimize the photometric loss in Eq.~\eqref{eq:inv}.
At time step $t$, we update the current reflectance map $x_{t}$ and the lighting parameters $\theta_t$ as:
\begin{align}
    x_{t-1}^{\prime} &= \frac{1}{\sqrt{\alpha_t}}\cdot\left(x_t-\frac{1-\alpha_t}{\sqrt{1-\bar{\alpha}_t}} \cdot\epsilon\left(x_t, t\right)\right)+\sigma_t\cdot z \label{eq:dps_sample} \\
    x_{t-1} &= x_{t-1}^{\prime} - \zeta_t \cdot\nabla_{x_t}\mathcal{L}_{pho}(\hat{x_t},\theta_t) \label{eq:dps_grad} \\
    \theta_{t-1} &= \theta_t -  \eta_t \cdot\nabla_{\theta_t}(\mathcal{L}_{pho}(\hat{x_t},\theta_t) + \mathcal{L}_{reg}(\theta_t))
\end{align}
Intuitively, we first denoises $x_t$ to a cleaner sample $x_{t-1}^{\prime}$ using Eq.~\eqref{eq:diffusion_reverse}. 
Then, we move $x_{t-1}^{\prime}$ towards the direction such that the clean estimation $\hat{x_t}$ minimizes the photometric loss $\mathcal{L}_{pho}$; we use $\zeta_t$ to control the step size.
Lastly, we update the lighting parameters $\theta_t$ to $\theta_{t-1}$ using gradient descent with a learning rate of $\eta_t$.
In addition to $\mathcal{L}_{pho}$, we apply a regularization term $\mathcal{L}_{reg}$ to the lighting parameters $\theta$ to ensure darker shading in regions with shadow-baking artifacts and spatial smoothness; see our \emph{supplementary material} for more details.

Note that when evaluating $\mathcal{L}_{pho}$, only the diffuse albedo component in $\hat{x_t}$ is used.
However, the specular albedo and detailed normal components in $x_{t-1}^{\prime}$ are also affected by $\nabla_{x_t}\mathcal{L}_{pho}(\hat{x_t},\theta_t)$ to follow the updation of the diffuse albedo component.
This way, the specular albedo and detailed normal map are implicitly constrained to align with the diffuse albedo map.
After the posterior sampling process, we read out the diffuse albedo map $A$, specular albedo map $S$, and detailed normal map $N_d$ from $x_0$.

\paragraph{Upsampling to 4K}
After solving the 1K-resolution reflectance maps, we use a super-resolution (SR) network $\mathcal{U}$~\cite{zhang2018rcan} to upsample them to 4K resolution; see our \emph{supplementary material} for more details of $\mathcal{U}$.
Compared to DoRA, which directly samples the 4K map from the diffusion model, our sample-then-SR design is significantly more efficient. 
Our method takes only 8 minutes using a 24 GB NVIDIA 4090, while DoRA requires 508 minutes.

\section{Experiments}
In this Section, we first introduce our implementation details in Section~\ref{sec:exp:id}.
We then evaluate core design choices in Section~\ref{sec:exp:eval} and compare our method to prior arts in Section~\ref{sec:exp:cmp}.
Next, we present results of our method on diverse subjects under different lighting conditions in Section~\ref{sec:exp:res} and discuss limitations in Section~\ref{sec:exp:limit}.
We strongly suggest the reader check our supplementary material and video in our \href{https://yxuhan.github.io/WildCap/index.html}{\textcolor{magenta}{project page}} for more experimental results.

\subsection{Implementation Details}\label{sec:exp:id}
We set the UV resolution to 1024, \emph{i.e.,} $H=W=1024$. 
In building the texture map $I_{UV}$, we minimize the combination of an LPIPS loss~\cite{zhang2018perceptual} and a gradient-space L1 loss between the rasterized images and $\{I^i\}_{i=1}^V$.
In our texel grid lighting model, we set the grid size $g=96$.
During optimization, we set the total sampling steps $T=1000$ and thus the actual sampling steps $T_{init}=600$.
We set $\zeta_t$ to 1. 
For $\eta_t$, we initialize it to 0.01 and apply an exponential learning rate decay scheduler to it.
All the experiments are conducted on a single 24GB NVIDIA RTX 4090.

\subsection{Evaluations}\label{sec:exp:eval}
In this Section, we evaluate several key design choices in our method, including the hybrid inverse rendering framework, the proposed texel grid lighting model, and the use of a diffusion prior for optimization.
In addition, we evaluate the effectiveness of skin tone control and provide a deeper analysis of TGL in the \emph{supplementary material}.

\paragraph{Evaluation on Hybrid Inverse Rendering}
We conduct a baseline, \emph{i.e., w/o hybrid}, where we directly apply our method to the raw captured images $\{I_{raw}^i\}_{i=1}^{V}$ instead of the predicted diffuse albedo images $\{I^i\}_{i=1}^V$.
As shown in Figure~\ref{Fig:cmp_ablat} (h), this baseline struggles to disentangle high-quality reflectance from faces with complex lighting effects such as specularity and shadows.
Our hybrid framework uses a data-driven method to remove most lighting effects, providing a good initialization for model-based optimization and thus achieving significantly better results than this baseline.

\begin{figure}[t]
    \centering
    \includegraphics[width=0.475\textwidth]{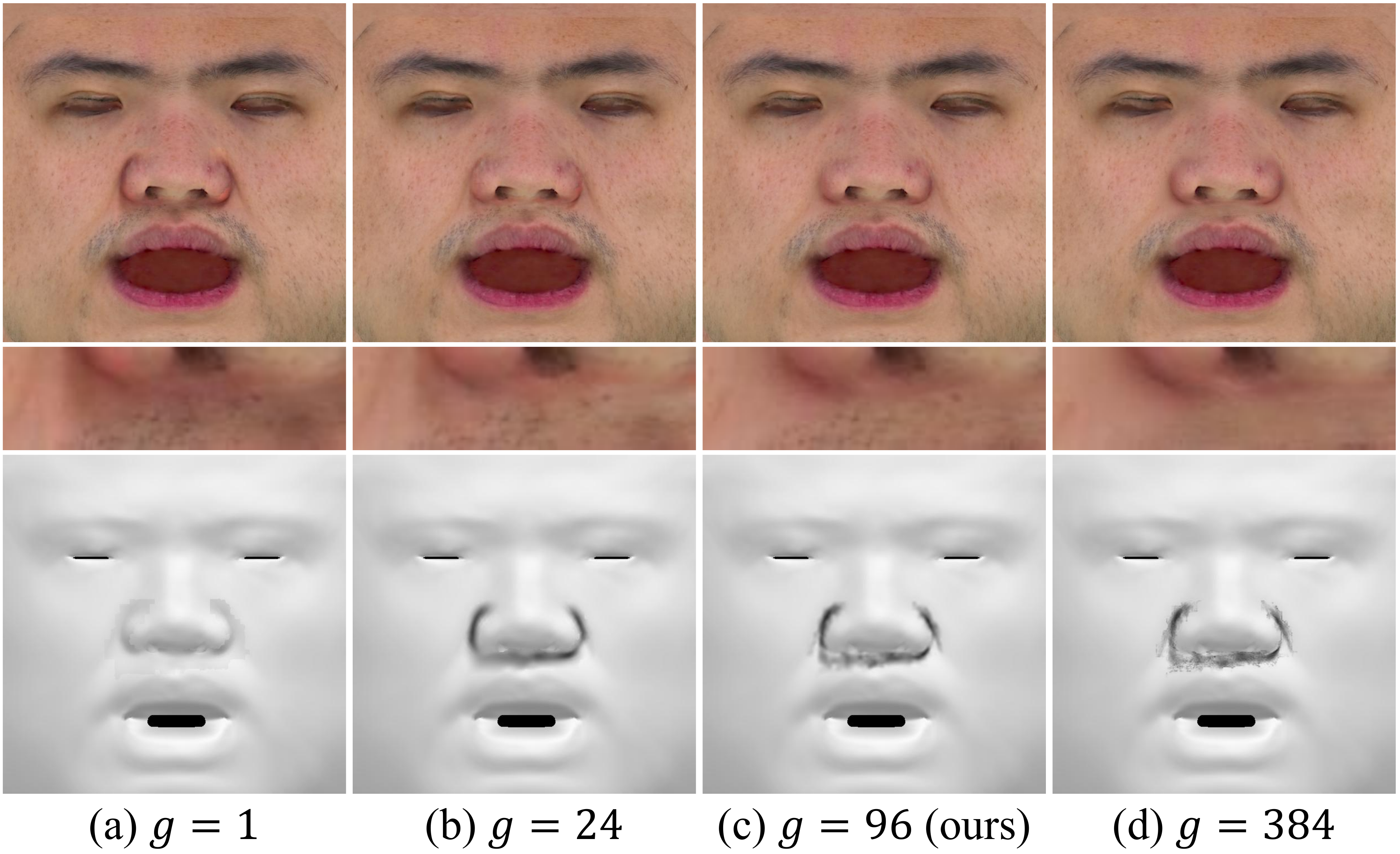}
    \caption{\textbf{Ablation study on the grid size $g$.} We show the reconstructed diffuse albedo map, close-up, and shading from the top row to the bottom row. }
    \label{Fig:ablat_grid_size}
\end{figure}

\paragraph{Evaluation on Texel Grid Lighting}
We conduct a baseline, \emph{i.e., w/o TGL}, where we apply a global SH lighting to model the lighting effects in $I_{UV}$.
As shown in Figure~\ref{Fig:cmp_ablat} (i), this baseline fails to explain baking artifacts in $I^i$, \emph{i.e.} Figure~\ref{Fig:cmp_ablat} (b), as lighting effects.
The reason is that the baking artifacts in the predicted diffuse albedo images are not produced by a physical light source in the real world.
Our lighting model can well remove artifacts in network predictions and produce a clean diffuse albedo map thanks to its strong expressive power.

We further evaluate the effect of the grid size $g$ in our texel grid lighting model.
As shown in Figure~\ref{Fig:ablat_grid_size}, a small grid size (\emph{e.g.} $g=1$ and $g=24$) is less expressive in representing baking artifacts as lighting effects.
On the other hand, large grid size, \emph{e.g.} $g=384$, tends to explain facial details into the lighting effects, resulting in an over-smooth texture.
We set $g=96$, achieving a good balance between removing the artifacts and reconstructing facial details.

\paragraph{Evaluation on Diffusion Prior}
We conduct a baseline, \emph{i.e., w/o prior}, where we enforce no regularization on the diffuse albedo map $A$ and directly optimize each texel using Adam~\cite{adam2014method}.
As shown in Figure~\ref{Fig:cmp_ablat} (j), this baseline produces severe artifacts because we have no guarantee of converging to a valid reflectance map illuminated by a dark light as we expect.
By solving the diffuse albedo in the valid distribution modeled by our diffusion prior $\epsilon$, we address this ill-posedness elegantly.

\begin{figure}[t]
    \centering
    \includegraphics[width=0.475\textwidth]{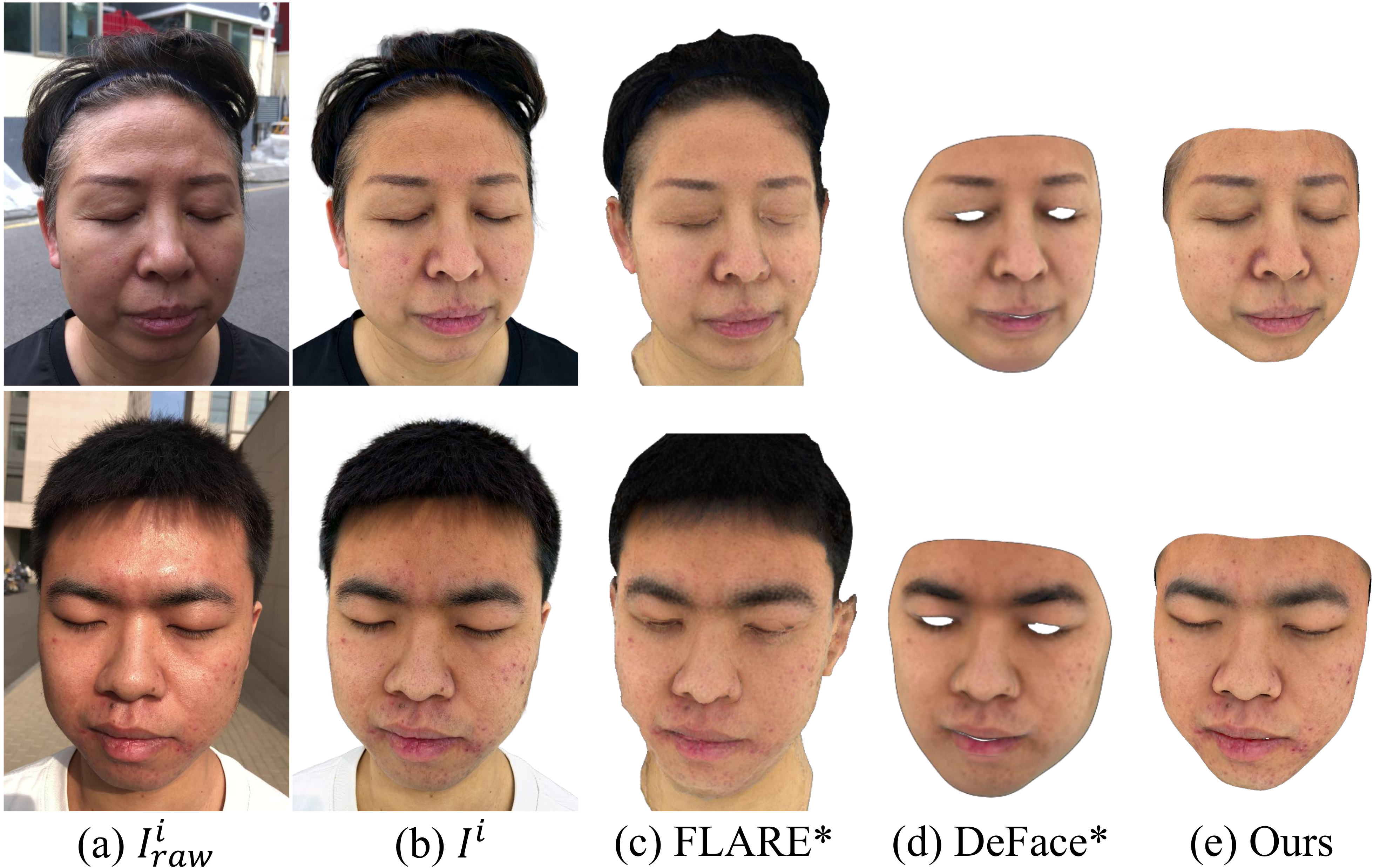}
    \caption{\textbf{Qualitative comparison on face reconstruction.} We present the face reconstruction results of FLARE*, DeFace*, and our method in (c), (d), and (e), respectively. We show the ground truth in (b) and the raw captured image in (a) for reference.}
    \label{Fig:cmp_recon}
\end{figure}

\begin{table}[t]
\centering
\small
\begin{tabular}{@{}lccc@{}}
\toprule
 & PSNR~$\uparrow$ & SSIM~\cite{wang2004image}~$\uparrow$ & LPIPS~\cite{zhang2018perceptual}~$\downarrow$  \\ \midrule
DeFace* & 22.20 & 0.9279 & 0.1192 \\
FLARE* & 27.81 & 0.9411 & 0.0929 \\ \midrule
 Ours & \textbf{28.79} & \textbf{	0.9520} & \textbf{0.0610}  \\ \bottomrule
\end{tabular}
\caption{\textbf{Quantitative comparison on face reconstruction.} The metric is averaged on $6$ subjects and computed on the same cropped facial skin region.}
\label{tab:cmp}
\end{table}

\subsection{Comparisons}\label{sec:exp:cmp}
In this Section, we first compare with in-the-wild methods using the same captured data.
We then compare our method with DoRA~\cite{han2025dora} on the same subjects but with different capture setups to evaluate the performance gap between our in-the-wild capture and the controllable capture.

\paragraph{Comparison to In-the-Wild Methods}
We consider DeFace~\cite{huanglearning} and FLARE~\cite{FLAME:SiggraphAsia2017} for comparison.
See our \emph{supplementary material} for a discussion of expected comparison results to closed-source works~\cite{xu2024monocular,rainer2023neural}.
DeFace takes a single-view image as input.
It optimizes a neural network with the facial reflectance maps to segment the face into regions, where each region is modeled with different SH lighting.
FLARE takes a monocular video as input.
It uses split-sum approximation~\cite{Munkberg_2022_CVPR} to model lighting and FLAME~\cite{FLAME:SiggraphAsia2017} with per-vertex displacement as facial geometry.
Note, DeFace and FLARE take the raw captured images $\{I_{raw}^i\}_{i=1}^v$ as input.
In addition, we construct two extra baselines, \emph{i.e.}, DeFace* and FLARE*.
Similar to our method, we feed them the predicted diffuse albedo images $\{I^i\}_{i=1}^v$ as input while leaving their other implementations unchanged.

As shown in Figure~\ref{Fig:cmp_ablat}, FLARE and DeFace fail to remove the complex lighting effects in the input.
That is because directly running inverse rendering on raw captured images is challenging and prone to local minima.
By augmenting with our hybrid inverse rendering framework, FLARE* and DeFace* produce better results.
However, they still keep most of the baking artifacts in network predictions.
A possible reason is the limited expressiveness of their lighting model.
FLARE* uses a split-sum lighting model, which fails to model non-physical lighting effects in network predictions.
Although the lighting model of DeFace* is conceptually similar to our method, their design choice restricts them to segment the face into a limited number of regions, such as 5 or 10, since each facial region is corresponded to a trainable network.
Thus, their expressiveness is limited compared to our method.
In Figure~\ref{Fig:cmp_ablat} (g), our method obtains the best diffuse albedo reconstruction results, with significantly fewer baking artifacts.

We further compare our method with DeFace* and FLARE* on face reconstruction.
In this experiment, all the methods take the predicted diffuse albedo images $\{I^i\}_{i=1}^V$ as input. 
Thus, we can compare the re-rendered images against $\{I^i\}_{i=1}^V$.
As shown in Figure~\ref{Fig:cmp_recon} and Table~\ref{tab:cmp}, our method obtains the best results.

\begin{figure}[t]
    \centering
    \includegraphics[width=0.475\textwidth]{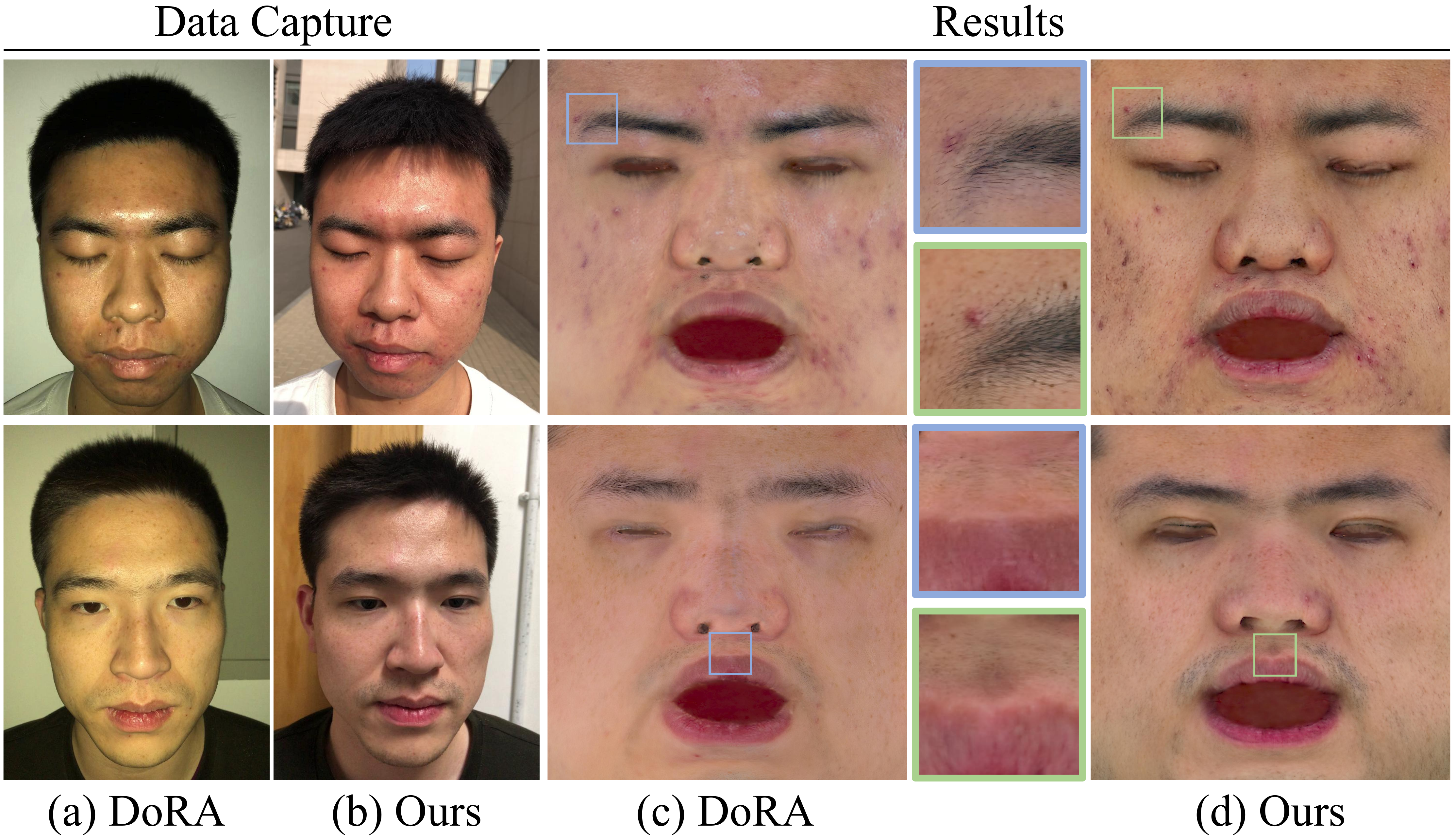}
    \caption{\textbf{Qualitative comparison with DoRA.} We capture a co-located (a) and in-the-wild (b) sequence for the same subject as the input to DoRA and our method, respectively. We then compare the reconstructed diffuse albedo map of DoRA (c) and our method (d).}
    \label{Fig:cmp_dora}
\end{figure}

\begin{figure*}[t]
    \centering
    \includegraphics[width=1\textwidth]{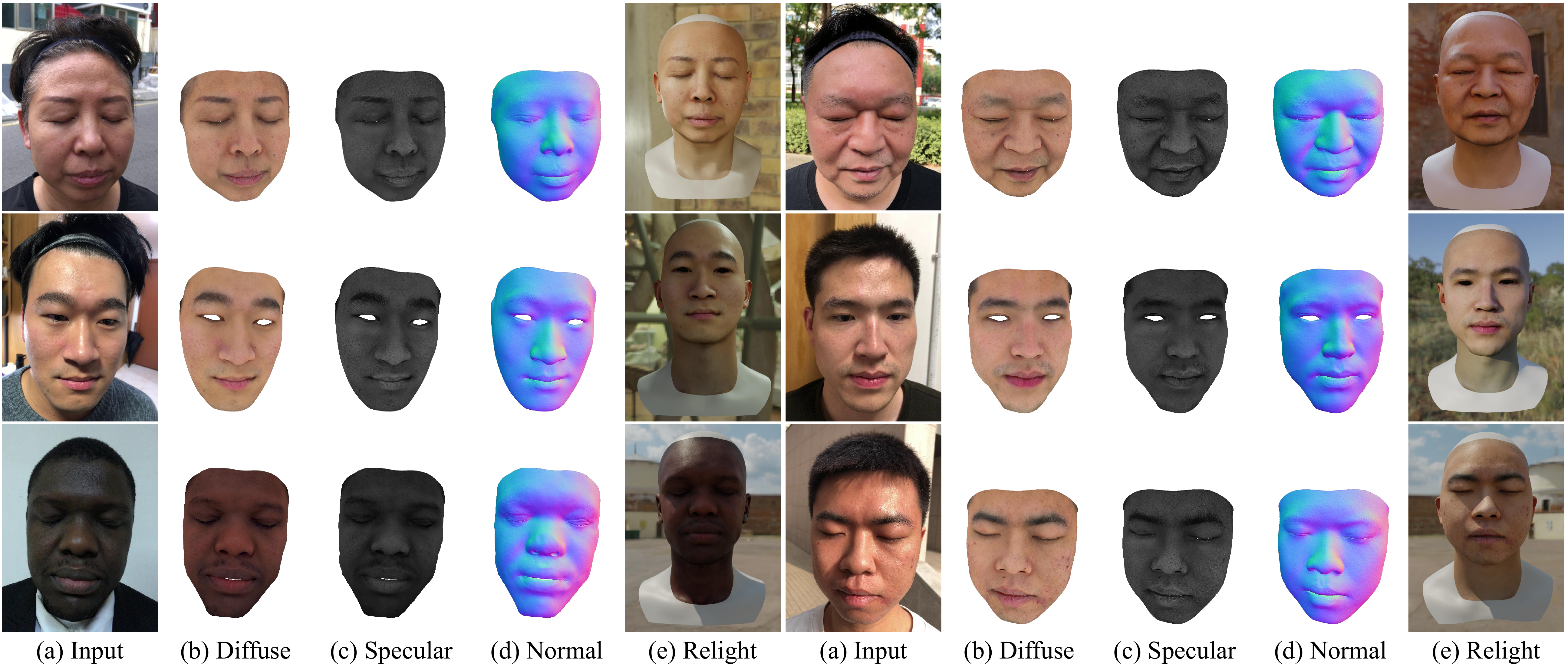}
    \caption{\textbf{Reflectance estimation and relighting results of our method on subjects captured in diverse environments}. }
    \label{Fig:results}
\end{figure*}

\paragraph{Comparison to DoRA}
We compare our method to DoRA~\cite{han2025dora}, a state-of-the-art method for low-cost facial appearance capture under controllable lighting.
We capture an extra co-located smartphone and flashlight video for the same subject.
Then, we feed the in-the-wild video to our method and the co-located video to DoRA for reconstruction.
As shown in Figure~\ref{Fig:cmp_dora}, our method demonstrates comparable quality to DoRA while significantly reducing capture cost.
In addition, our method can better preserve person-specific facial traits such as nevus, achieving high-fidelity results.
These facial traits are recovered by the texture building step thanks to the LPIPS loss.
DoRA fails to recover these details because its L2 photometric loss tends to average out person-specific facial traits due to inaccuracy in geometry reconstruction and camera calibration.
In our pilot experiments, we find that simply adding an LPIPS term to the posterior sampling process is brittle and hard to tune; our strategy effectively introduces the LPIPS loss into reflectance estimation via a robust texture building process. 

\subsection{More Results}\label{sec:exp:res}
We present the results of our method on diverse subjects in Figure~\ref{Fig:results}.
The videos are captured in diverse environments, including indoor and outdoor.
Although trained on only 48 Light Stage scans, our method generalizes well to unseen people, achieving high-quality reflectance estimation and relighting results; similar observations are also found in DoRA~\cite{han2025dora}.
We believe the reasons are two-fold.
On the one hand, our patch-level design improves generalization.
On the other hand, the diffusion posterior sampling technique is powerful to steer the diffusion model to reconstruct the signal.
In addition, our method can reconstruct high-quality facial reflectance, including diffuse albedo, specular albedo, and detailed normal, leading to photo-realistic renderings in new environments.

\subsection{Limitations and Discussions}\label{sec:exp:limit}
Firstly, our method relies on SwitchLight~\cite{kim2024switchlight} for preprocessing, which is a property model with only an API available.
Secondly, our automatic shadow-detection method relies on DiFaReli~\cite{ponglertnapakorn2023difareli}, which is slow due to its iterative diffusion sampling and has room for improvement.
Training a network for face-delighting with confidence estimation for shadow regions using the recently released FaceOLAT dataset~\cite{rao20253dpr} to replace SwitchLight and DiFaReli is an important future direction.

\section{Conclusion}
We propose WildCap for facial albedo capture from smartphone video recorded in the wild.
To achieve this, we design a hybrid inverse rendering method.
Our key idea is to use a robust data-driven method, \emph{i.e.}, SwitchLight, to convert the in-the-wild capture to a more constrained case.
Then, we apply a model-based optimization to explain the baking artifacts in the network predictions as lighting effects.
To model non-physical lighting effects in network predictions, we propose a novel texel grid lighting model.
Combined with the patch-level diffusion prior, our method achieves high-quality facial albedo estimation, filling the quality gap between in-the-wild methods and methods with controllable recordings by a large margin.

\section*{Acknowledgement}
This work was supported by the NSFC (No.62561160115). This work was also supported by THUIBCS, Tsinghua University, and BLBCI, Beijing Municipal Education Commission. Feng Xu is the corresponding author.

\appendix

\section{More Implementation Details}

\subsection{Shadow Detection}
As mentioned in the main paper, our method only requires a coarse mask $M$ to indicate the baking artifacts in $I_{UV}$.
Thus, obtaining $M$ is low-cost and easy.
In the following, we propose two methods to obtain $M$, one is manual, the other is fully automatic.

\paragraph{Manual Method}
For the manual method, we open $I_{UV}$ in Photoshop and use the Polygonal Lasso Tool to mark the facial regions containing baking artifacts.
This step is easy and only requires a few mouse clicks.

\begin{figure}[t]
    \centering
    \includegraphics[width=0.475\textwidth]{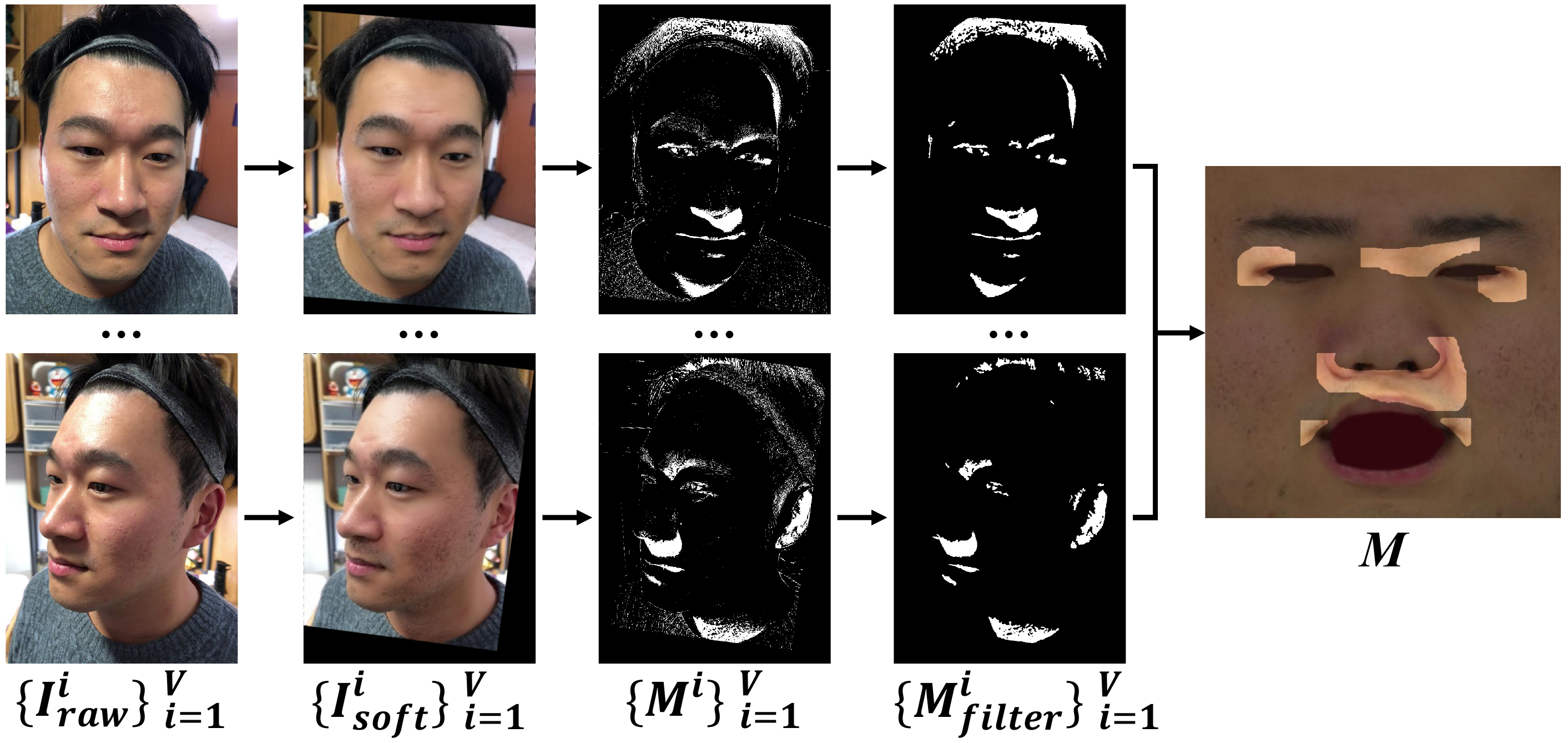}
    \caption{Pipeline of the proposed automatic method to obtain the shadow mask $M$.}
    \label{Fig:auto_shadow_mask}
\end{figure}

\paragraph{Automatic Method}
We also develop a fully automatic method to obtain $M$ as shown in Figure~\ref{Fig:auto_shadow_mask}.
Specifically, we detect shadow regions in the raw images $\{I_{raw}^i\}_{i=1}^V$, and then lift these image-space predictions into the UV space to obtain $M$.
The rationale is to use shadow as a proxy to locate baking artifacts.

To detect shadow regions in $\{I_{raw}^i\}_{i=1}^V$, we adopt an existing work, \emph{i.e.}, DiFaReli~\cite{ponglertnapakorn2023difareli}.
Following DiFaReli++~\cite{ponglertnapakorn2025difarelidiffusionfacerelighting}, we use DiFaReli to soften the shadows in $\{I_{raw}^i\}_{i=1}^V$.
We denote the processed images as $\{I_{soft}^i\}_{i=1}^V$.
Then, we compute the shadow mask $M^i$ by thresholding the color difference between $I_{raw}^i$ and $I_{soft}^i$.
We further apply a median filter to $M^i$ and remove connected areas smaller than a threshold; we denote the processed per-view shadow mask as $M^i_{filter}$
Next, we lift $\{M^i_{filter}\}_{i=1}^V$ to the UV space to obtain $M$; we also dilate $M$ to some extent to ensure it includes all the baking artifacts.
All the hyperparameters, \emph{e.g.}, thresholds and kernel sizes, are shared across different subjects.

\begin{figure}[t]
    \centering
    \includegraphics[width=0.475\textwidth]{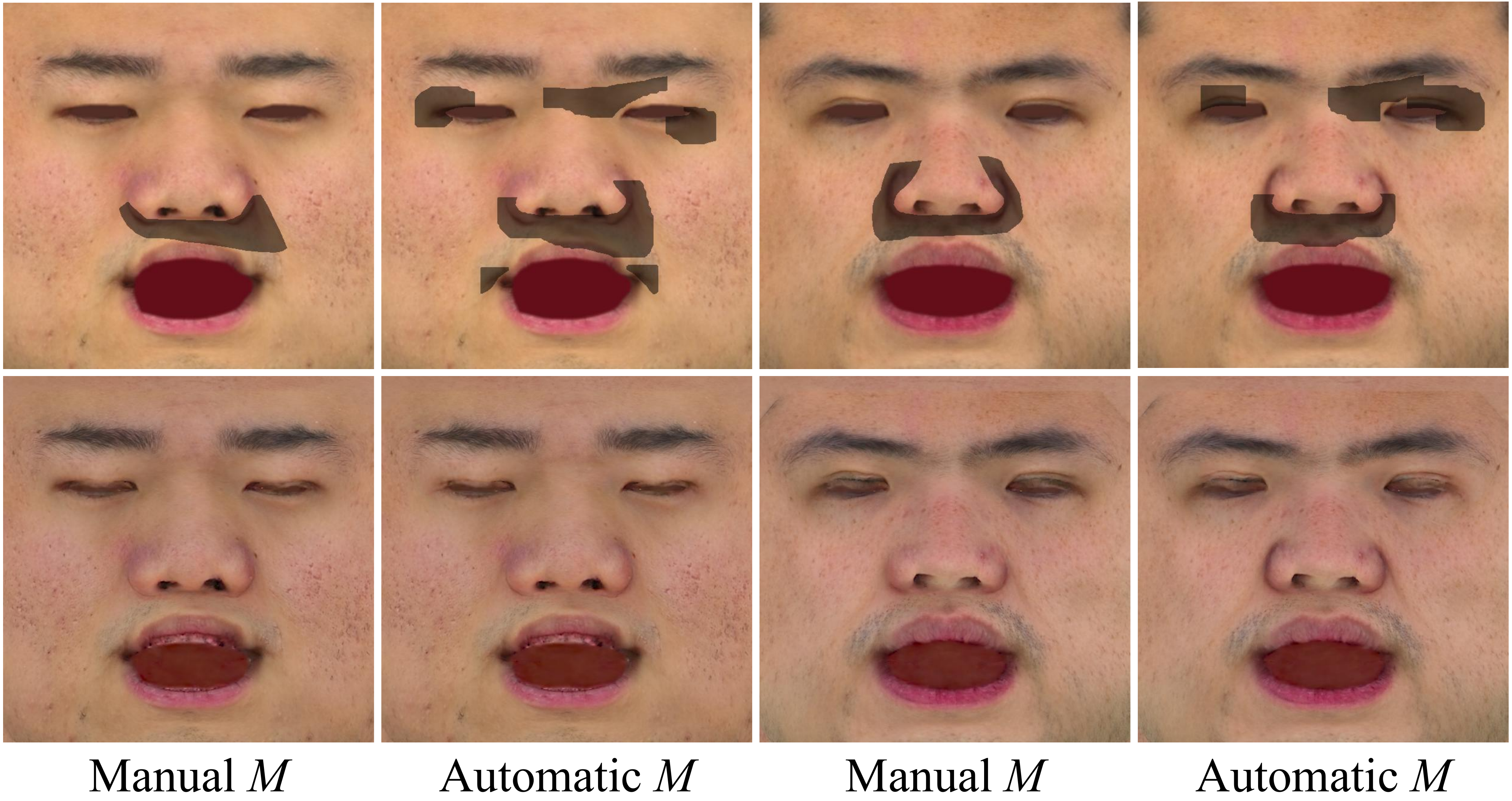}
    \caption{Comparison of the manual and automatic shadow mask $M$. We visualize the shadow mask in the first row and show the reconstructed diffuse albedo map in the second row. }
    \label{Fig:cmp_shadow_mask}
\end{figure}

\paragraph{Comparison of the Two Methods}
As shown in Figure~\ref{Fig:cmp_shadow_mask}, the automatic and manual methods reconstruct diffuse albedo maps in similar quality.
Since the goal of our automatic method is to detect shadow regions as a proxy for baking artifacts, it also includes regions around eyes in $M$.
However, we find that SwitchLight produces negligible baking artifacts around the eyes in the 2 cases shown in Figure~\ref{Fig:cmp_shadow_mask}, thus we do not mark them in the manual method.
In addition, we notice that the automatic method fails to detect the baked ambient occlusion effects on the side nose, as shown in the rightmost column.
To ensure the highest quality, we use the manual method by default.
We leave training a portrait-delighting network with shadow removal confidence as our future work.

\subsection{Light Stage Dataset}
Our Light Stage dataset for training the diffusion prior is the same as that used in DoRA~\cite{han2025dora}.
The dataset contains 6 Asians (2 males and 4 females), 9 African Americans (5 males and 4 females), and 33 Caucasians (17 males and 16 females).
Please refer to DoRA for details on processing the dataset.

\subsection{Lighting Regularization}
As mentioned in the main paper, during optimization, we add a regularization term $\mathcal{L}_{reg}$ to our lighting model $\Gamma_\theta$:
\begin{equation}
    \mathcal{L}_{reg} = 0.1\cdot\mathcal{L}_{TV} + \mathcal{L}_{neg}
\end{equation}
We apply a total variation regularization $\mathcal{L}_{TV}$ to constrain the spatial smoothness of the actual lighting parameters $\gamma$:
\begin{equation}
    \mathcal{L}_{TV} = \sum_{u,v} ||\gamma_{u,v}-\gamma_{u,v-1}||_2^2 + ||\gamma_{u,v}-\gamma_{u-1,v}||_2^2
\end{equation}
We apply a negative shading regularization $\mathcal{L}_{neg}$ to constrain the shading of $\gamma^V$ to be negative:
\begin{equation}
    \mathcal{L}_{neg} = \sum_{u,v} \max(0,s^V_{u,v})^2
\end{equation}
Here, $s^V_{u,v}$ is the shading of $\gamma^V$ at UV location $(u,v)$.
The rationale of $\mathcal{L}_{neg}$ is that we expect baking artifacts to be explained as a clean diffuse albedo map illuminated by local dark lights.

\subsection{Super-Resolution Network}
We adopt RCAN~\cite{zhang2018rcan} as our super-resolution network $\mathcal{U}$ to upsample the 1K resolution reflectance maps into 4K.
Similar to previous works~\cite{lattas2020avatarme}, we train $\mathcal{U}$ at the patch level.
At inference time, we directly send a 1K-resolution reflectance map to $\mathcal{U}$.
During training, we cropped paired reflectance patches from the 1K and 4K versions of the scan. 
The patch size is set to $48\times48$, and $\mathcal{U}$ is trained to upsample it to $192\times192$.
We also modify the number of input and output channels of RCAN to 7 to support upsampling the concatenated diffuse albedo, specular albedo, and detailed normal map simultaneously. 

\section{More Experiments}

\begin{figure}[t]
    \centering
    \includegraphics[width=0.475\textwidth]{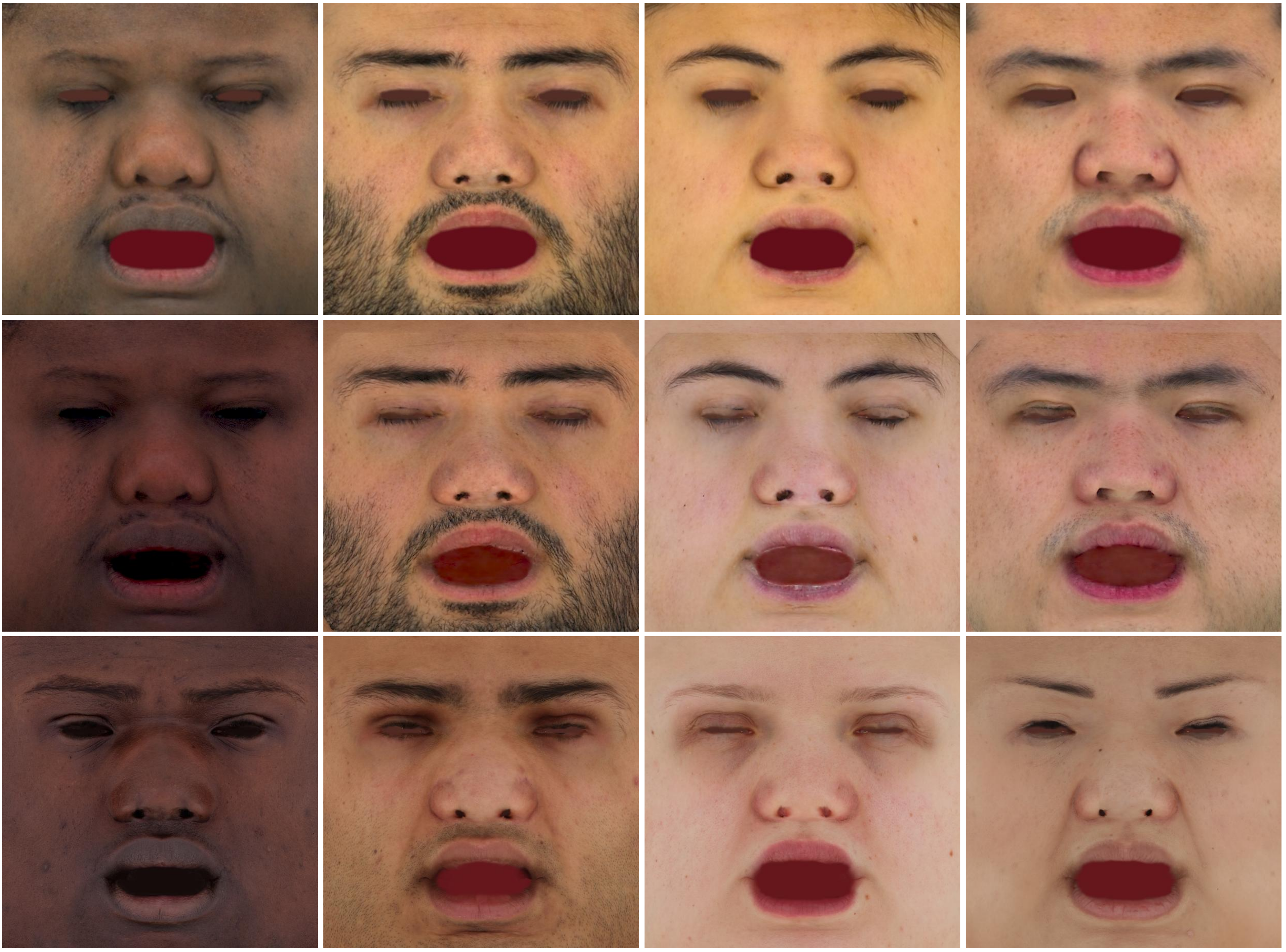}
    \caption{Evaluation on skin tone control. We show the texture map $I_{UV}$, the reconstructed diffuse albedo map $A$, and the initialization $x_0^{ref}$ from the top row to the bottom row. }
    \label{Fig:eval_skin_tone}
\end{figure}

\subsection{Evaluation on Skin Tone Control}
Recall that in our method, we control the skin tone via initialization.
Specifically, we set the sampling start point $x_{T_{init}}$ as the noised version of a Light Stage scan $x_0^{ref}$ whose skin tone is similar to the provided one. 
We also modify the diffuse albedo component of $x_0^{ref}$ using the color-matching transform to better align with the provided skin tone.
At the same time, we initialize the lighting so that the shaded initial reflectance map $x_0^{ref}$ has a consistent color tone as $I_{UV}$.
As shown in Figure~\ref{Fig:eval_skin_tone}, our strategy can effectively control the skin tone of the solved diffuse albedo maps (2nd row) to match the initialization $x_0^{ref}$ (3rd row).

\begin{figure*}[p!]
    \centering
    \includegraphics[width=0.95\textwidth]{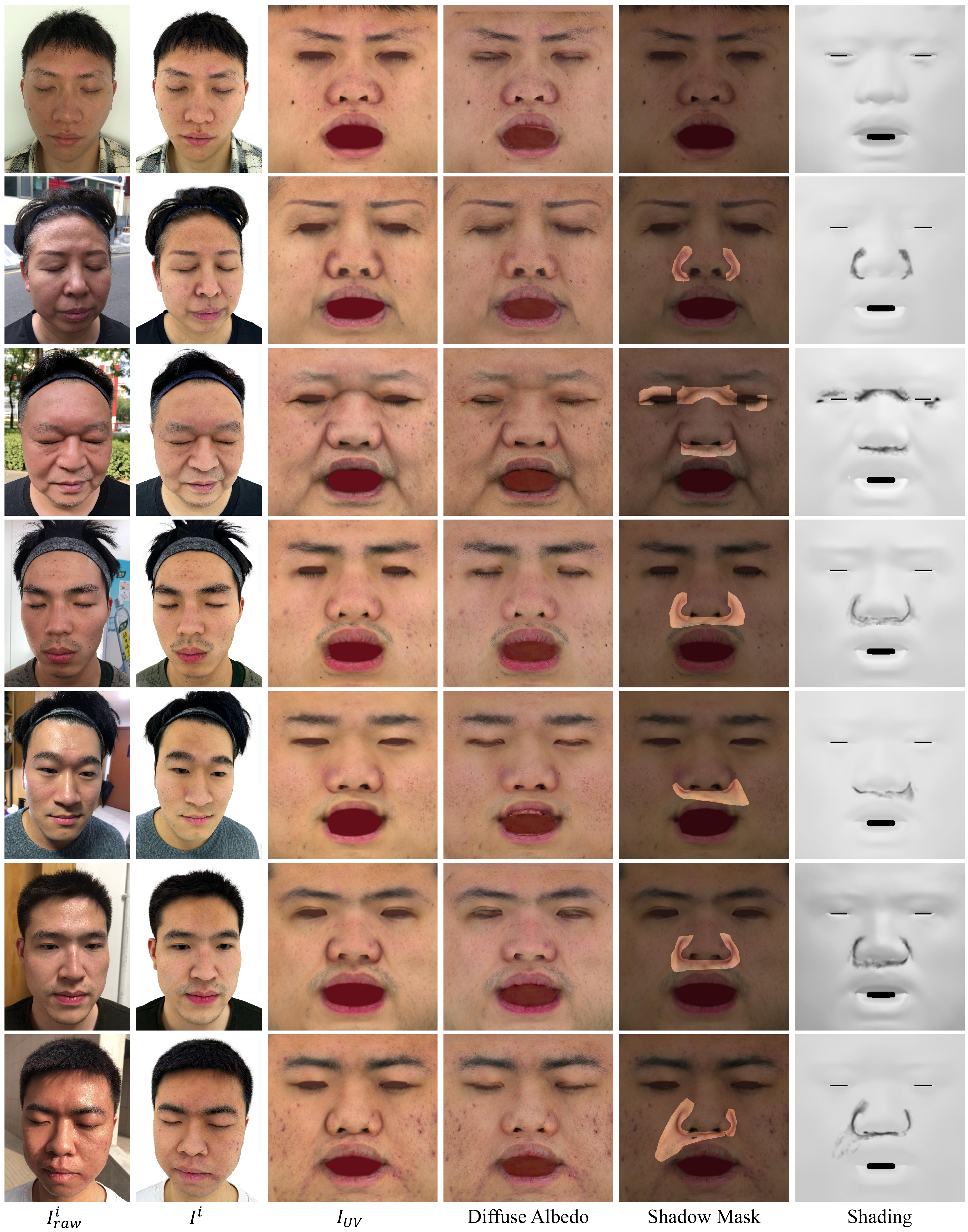}
    \caption{Evaluation of our method on various in-the-wild captures. From left to right, we show a raw captured image $I_{raw}^i$, the predicted diffuse albedo image $I^i$ by SwitchLight, the texture $I_{UV}$, the reconstructed diffuse albedo map, the shadow mask used to modulate our texel grid lighting model, and the shading.}
    \label{Fig:eval_switchlight}
\end{figure*}

\subsection{Baking Artifacts of SwitchLight}
Since one of our core contributions is to explain SwitchLight's baking artifacts as lighting effects, a natural question is, when will SwitchLight produce these artifacts?
In Figure~\ref{Fig:eval_switchlight}, we comprehensively test our method on diverse in-the-wild cases, ranging from simple cases captured under low-frequency lighting to hard cases with apparent shadow and specularity appearing on the face.

From Figure~\ref{Fig:eval_switchlight}, we find that SwitchLight performs quite well in easy cases with low-frequency lighting, such as the first two rows.
As the scene illumination becomes high-frequency, shadows and specularity appear on the face.
We empirically find that SwitchLight works well in removing specularity, but struggles in shadows.
However, shadows are ubiquitous in everyday captures. 
For example, both the sun and the roof light bulb would cast shadows on the face.
This drawback prevents SwitchLight from becoming an ideal method for facial albedo capture in the wild.
Fortunately, thanks to our model-based optimization, we successfully explain the shadow-baking artifacts as a clean diffuse albedo illuminated by a dark shading.

\begin{figure*}[h]
    \centering
    \includegraphics[width=\textwidth]{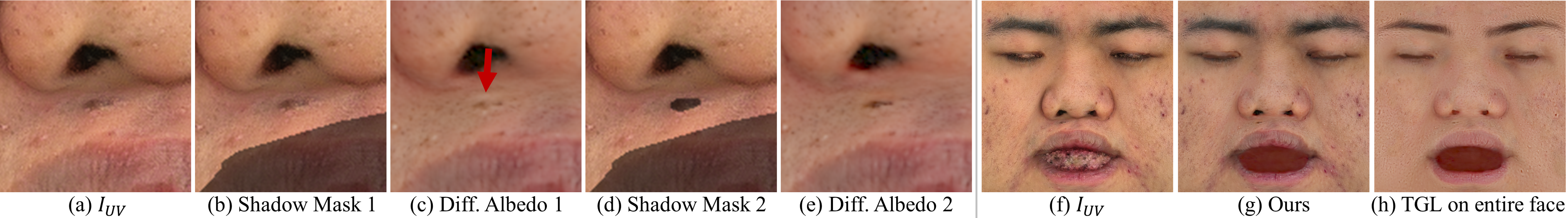}
    \caption{A deeper analysis of the Texel Grid Lighting.}
    \label{Fig:deep_tgl}
\end{figure*}

\subsection{Deeper Analysis of Texel Grid Lighting}
As shown in Figure 5 of the main paper, there is a trade-off controlled by the grid size: a larger grid (with smaller $g$) can better preserve details, while a smaller grid (with larger $g$) can produce a cleaner albedo due to increased representation capacity.
Here, we test on a more challenging case to highlight the potential loss of real texture details due to the high-capacity representation of TGL.

In an extreme case shown in Figure~\ref{Fig:deep_tgl} where \emph{the nevus is located in the shadow region and its size is coincidentally close to the grid size} (a), our method ($g=96$) somehow absorbs it into the lighting (b)(c).
This is because the small grid lacks global information to distinguish nevus from baking shadow.
To address this while preserving the expressive power of the small grid, we can exclude the nevus from the shadow mask (d)(e).
This aligns with our design: we apply TGL only to the shadow region (g), rather than the entire face (h).

\begin{table}[t]
\centering
\begin{tabular}{@{}lccc@{}}
\toprule
 & PSNR~$\uparrow$ & SSIM~\cite{wang2004image}~$\uparrow$ & LPIPS~\cite{zhang2018perceptual}~$\downarrow$  \\ \midrule
DeFace* & 28.43 & 0.9791 & 0.0826 \\
FLARE* & 22.48 & 0.9742 & 0.0571 \\ \midrule
 Ours & \textbf{28.71} & \textbf{0.9802} & \textbf{0.0388}  \\ \bottomrule
\end{tabular}
\caption{Quantitative comparison on diffuse albedo reconstruction. We compare our method with DeFace* and FLARE* using the scan of Digital Emily. The metric is computed on the same cropped facial skin region.}
\label{tab:cmp_emily}
\end{table}

\subsection{Quantitative Comparison on Synthetic Data}
We conduct a quantitative comparison of diffuse albedo reconstruction using the Digital Emily project~\cite{alexander2009digital}.
We render multi-view images from the scan and then run each method (Ours, DeFace*~\cite{huanglearning}, and FLARE*~\cite{bharadwaj2023flare}).
We compare the reconstructed diffuse albedo and GT in image space.
As shown in Table~\ref{tab:cmp_emily}, our method obtains the best metrics.

\begin{figure*}[t]
    \centering
    \includegraphics[width=1\textwidth]{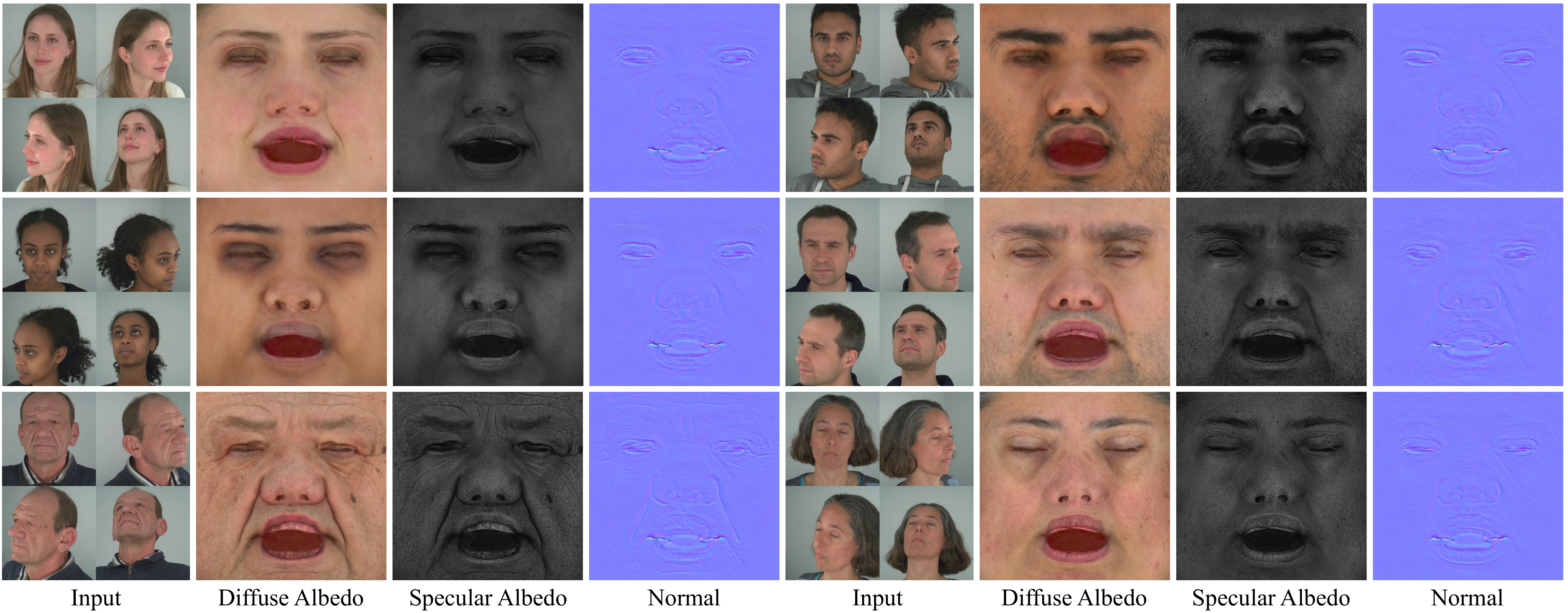}
    \caption{Results of our method on the NeRSemble~\cite{kirschstein2023nersemble} dataset (4 of 16 captured images are shown above).}
    \label{Fig:nersemble}
\end{figure*}

\subsection{Results on Studio-Captured Dataset}
Our method can also be applied to studio-captured multi-view face datasets, like NeRSemble~\cite{kirschstein2023nersemble} and Ava256~\cite{martinez2024codec}.
We show some results on NeRSemble in Figure~\ref{Fig:nersemble}.
Compared to in-the-wild videos captured by a smartphone camera, these studio-captured datasets are less challenging. 
The reason is that the lighting conditions in these studio-captured datasets are low-frequency.
For example, Ava256 uses uniform white light to capture the data, and the captured images are almost shadow-free.
We believe using our method to create an open-sourced, large-scale Light Stage dataset from existing studio-captured datasets is a valuable future direction.

\begin{figure*}[t]
    \centering
    \includegraphics[width=\textwidth]{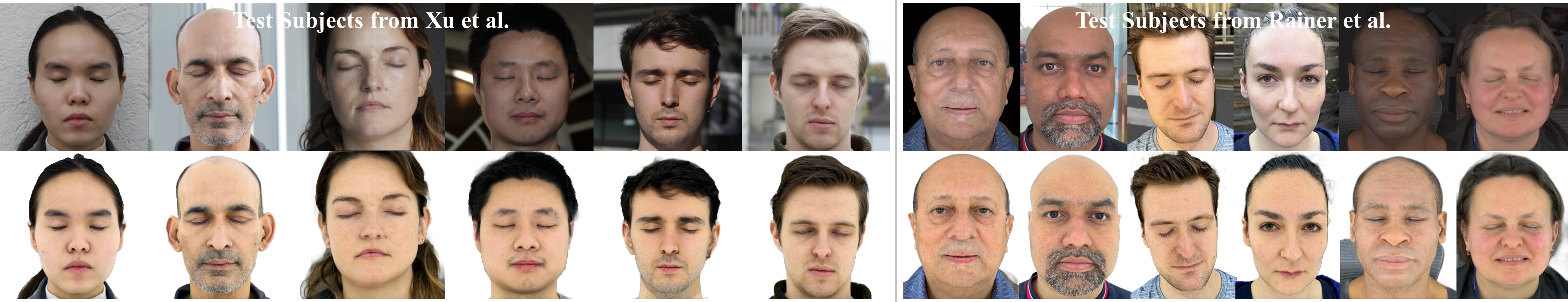}
    \caption{Switchlight's prediction (the second row) of test subjects in \citet{xu2024monocular} and \citet{rainer2023neural} (the first row).}
    \label{Fig:contribution}
\end{figure*}

\subsection{Position and Contribution of WildCap}
Despite previous closed-source works, such as \citet{xu2024monocular} and \citet{rainer2023neural}, proposing to capture appearance from multi-view images, \emph{we emphasize that we consider a more challenging and practical problem}. 

As shown in Figure~\ref{Fig:contribution}, the test subjects in \citet{xu2024monocular} and \citet{rainer2023neural} have \emph{little or moderate} cast shadow (1st row), which can already be well resolved by SwitchLight~\cite{kim2024switchlight} (2nd row) or our baseline method \emph{w/o TGL}.
However, we test on subjects with \emph{strong} cast shadow (\emph{e.g.}, the last 3 rows in Figure~\ref{Fig:eval_switchlight}), where SwitchLight leaves apparent baking artifacts.
We emphasize that cast shadow from indoor lighting or the sun is ubiquitous in the real world. 
Although we process more challenging data, our diffuse albedo maps have fewer artifacts and more details than Figure 1 of \citet{xu2024monocular} and Figure 5 of \citet{rainer2023neural}. 

Thus, in addition to technical novelty, we convey to the community that low-cost techniques can handle challenging cases with strong cast shadows, which is \emph{a new effect and a large step} in this field.

\begin{figure}[t]
    \centering
    \includegraphics[width=0.475\textwidth]{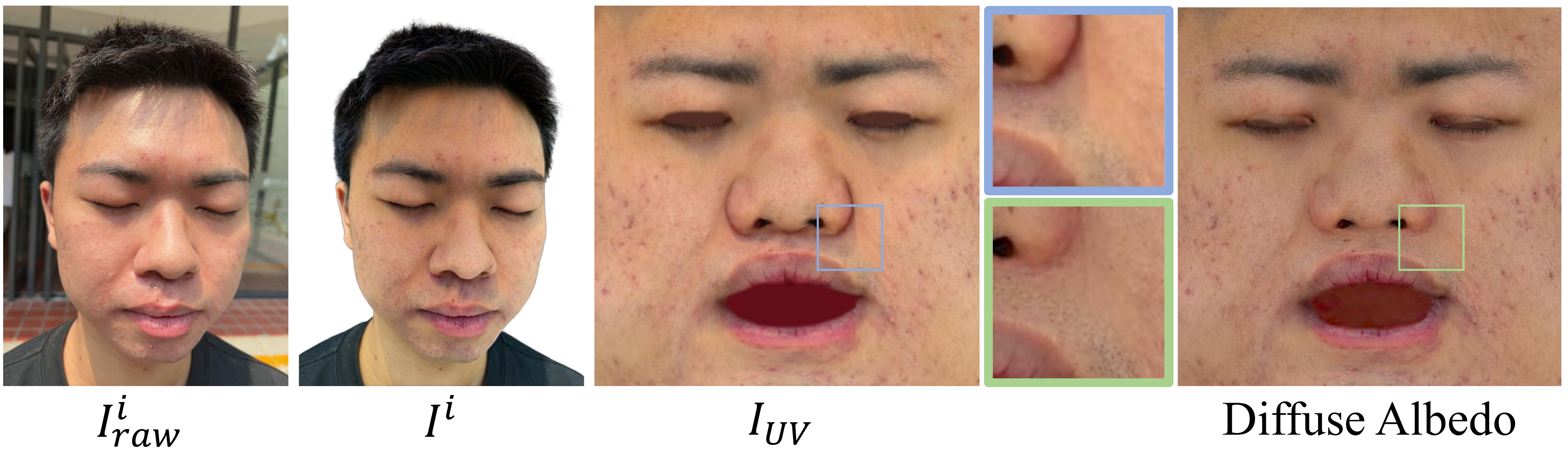}
    \caption{Failure case of our method. From left to right, we show the raw captured image $I_{raw}^i$, the predicted diffuse albedo image $I^i$ by SwitchLight, the texture map $I_{UV}$, close-ups, and the reconstructed diffuse albedo map.}
    \label{Fig:failure}
\end{figure}

\subsection{Failure Case of WildCap}
Since our lighting representation is continuous, our method does not perform well when sharp shadow boundaries appear in SwitchLight's prediction.
As shown in Figure~\ref{Fig:failure}, we test on a challenging case where the video is captured at noon under the sun.
Even after being delighted by SwitchLight, there are still very sharp shadow boundaries on the face.
Although our method obtains significantly better results, it still cannot totally remove these sharp boundaries.
To address this, we leave training an improved portrait-delighting network as our future work.

{
    \small
    \bibliographystyle{ieeenat_fullname}
    \bibliography{main}
}


\end{document}